\documentclass{article}

\usepackage[nonatbib]{nips_2018}

\usepackage[utf8]{inputenc} 
\usepackage[T1]{fontenc}    
\usepackage{hyperref}       
\usepackage{url}            
\usepackage{enumerate}
\usepackage{amsmath}
\usepackage{dcolumn}
\usepackage{booktabs}       
\usepackage{amsfonts}       
\usepackage{nicefrac}       
\usepackage{microtype}      
\usepackage{float}
\usepackage{graphicx,psfrag,epsf}
\usepackage{systeme}
\usepackage[linesnumbered,ruled,vlined]{algorithm2e}
\makeatletter
\renewcommand{\@algocf@capt@plain}{above}
\makeatother

\usepackage{verbatim}

\title{Neural Networks for Lorenz Map Prediction: A Trip Through Time.} 
\author{Denisa ~Roberts \\
 AI SpaceTime, NYC, NY\\
  \texttt{d.roberts@aispacetime.org} \\
    }

\begin{document}

\maketitle

\begin{abstract}

In this article the Lorenz dynamical system is revived and revisited and the current state of the art results for one step ahead forecasting for the Lorenz trajectories are published. Multitask learning is shown to help learning the hard to learn z trajectory. The article is a reflection upon the evolution of neural networks with respect to the prediction performance on this canonical task.

\end{abstract}

\section{Introduction and Historical Context}

Time series prediction via nonlinear models goes back in time. 
In his work \textit{``Cybernetics: or Control and Communication in the Animal and the Machine"}, Norbert Wiener poses the question in the second edition to his original 1948 book:

 \textit{``Can man-made machines learn and can
they reproduce themselves? We shall try to show in this chapter that in fact they can learn and can reproduce themselves, and we shall give an account other technique needed for both these activities.
The simpler of these two processes is that of learning, and it is there that the technical development has gone furthest. I shall talk here particularly of the learning of game-playing machines which enables them to improve the strategy and tactics of their performance by experience. $\ldots$ The statistical studies necessary to use a long past for a determination of the policy to be adopted in view of the short past are highly non-linear $\ldots$ In general, a learning
machine operates by non-linear feedback.''} 

In his original edition of the book he gives lengthy treatment to topics such as non-linear system learning, time series, information and communication, feedback and oscillation and the relation to learning and self-reproducing machines (first edition published in 1948 \cite{wiener1948cybernetics} and second edition was published in 1965 \cite{wiener1965cybernetics}). 

The first wave of artificial neural nets research overlapped with the moniker ``Cybernetics'', as discussed in the introduction to the ``Deep Learning'' textbook by Goodfellow et al. \cite{goodfellow2016deep}, followed by the next wave populated by terms such as connectionism and artificial neural networks, and then finally followed by the current wave of deep learning. Artificial neural networks were typically studied in relation to non-linear dynamics as discussed by Wiener ever since the 1940's. Of course the cybernetic discourse started much earlier than that, perhaps with Ada Lovelace, perhaps even earlier (see Wiener's comments on Goethe's ``Sorcerer's Apprentice''). 
 Many real world systems have been modeled (or attempted to) via time series with non-linear dynamics, such as the economic machine (macro and microeconomic dynamics, financial markets etc.) and physical dynamics with chaotic trajectories. A canonical example used throughout history to illustrate non-linear dynamics and chaos had been the Lorenz map and we revisit this example in a comparison study of prediction performance at different stages of neural network research.

As an aside, dynamical systems have been employed to characterize the training of neural networks themselves. Pascanu et. al draws a parallel in \cite{pascanu2013difficulty} between the training of recurrent neural networks and the problem of vanishing and exploding gradient. More specifically, the convergence of the training is characterized as a dynamical system with asymptotic behavior described by one of several possible different attractor states depending on the starting conditions and the model can be found in a chaotic regime as well.

The contributions of this article are:
\begin{itemize}
\item provide historical context and a review of the Lorenz task
\item description of several architectures mentioned in literature at different points along the neural network for time series prediction history with Lorenz reported results
\item description of modifications that lead to improved results
\item comparison of results for conditional and unconditional and with different sampling schemes
\item discussion of sensitivity to data distribution change
\item end to end \href{https://github.com/D-Roberts/gluon-spaceTime/tree/master/LorenzMap}{code} to replicate results.

\end{itemize}

\section{Nonlinear Dynamics and The Lorenz Equations}

Strogatz \cite{strogatz2018nonlinear} begins the study of chaos in his textbook on nonlinear dynamics with a discussion of the Lorenz equations. The Lorenz system of ordinary differential equations is \cite{lorenz1963deterministic}.

\begin{align*}
\dot{x} = \sigma \left( y - x \right) \\
\dot{y} = rx - y - xz \\
\dot{z} = xy - bz 
\end{align*}

with $\sigma > 0$ as a parameter called the Prandtl number, $r > 0$ is the Rayleigh number, and $b > 0$ is another nameless parameter. Lorenz derived this three dimensional system from a model of atmosphere, but the equations can arise in models of laser, dynamos and the motion of waterwheels.

\begin{figure}[!htb]
    \centering
    \begin{minipage}{0.5\textwidth}
        \centering
        \includegraphics[width=\textwidth]{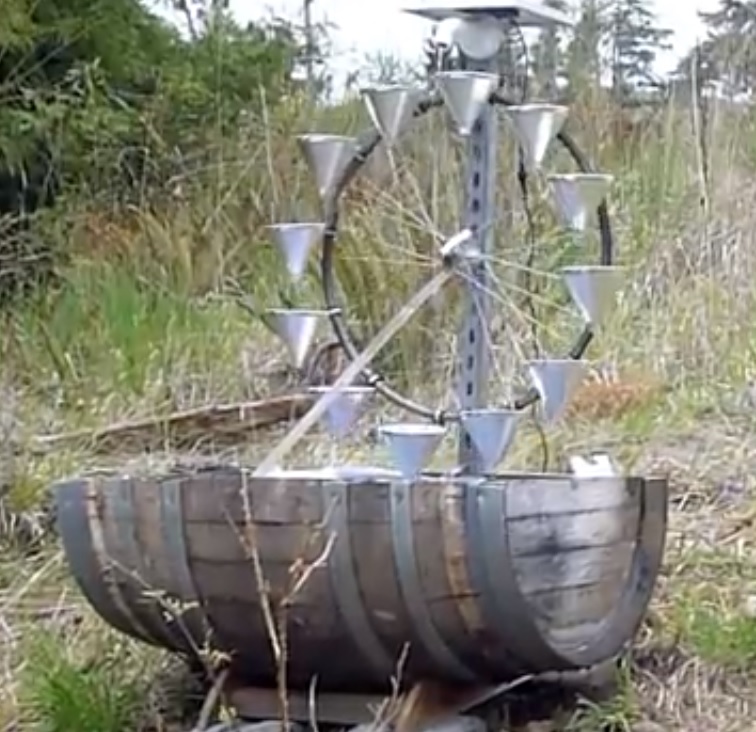}
    \end{minipage}
       \caption{Homemade Lorenz Water Wheel}
\end{figure}

In a simplified setup, the wheel has paper cups suspended from its rims. Water is poured from the top. Eventually the wheel starts turning, with rotation in either direction equally possible. By increasing the water flow, the steady rotation can be destabilized, and the motion can become chaotic rotating in one direction for a few turns, then changing directions, appearing to have an erratic dynamic. The chaotic behavior depends on initial conditions and the Lorenz equations had been used to model these dynamics. The nonlinear relationship between the pairwise series in the Lorenz map can best be seen in plots. When plotted in three dimensions, the trajectories give rise to a strange attractor. In two dimension, the famous Lorenz butterfly arises (when plotting z trajectory against x). The other two pairwise plots are given in Figure 2 as well.

\begin{figure}[htp]

\centering
\includegraphics[width=.3\textwidth]{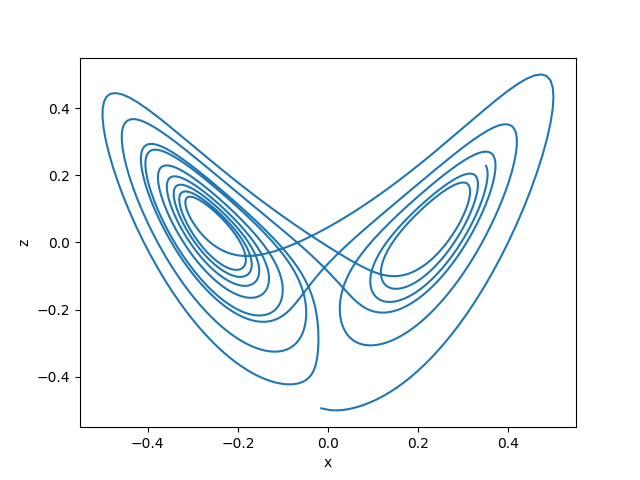}\hfill
\includegraphics[width=.3\textwidth]{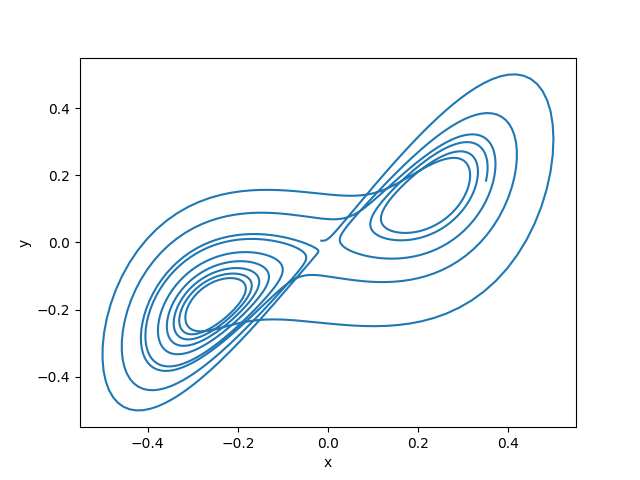}\hfill
\includegraphics[width=.3\textwidth]{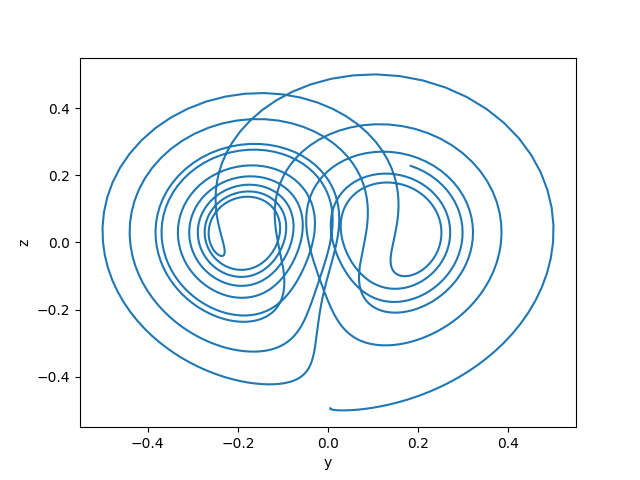}

\caption{Lorenz map pairwise trajectories}

\end{figure}

Not dissimilar to neural networks learning, the Lorenz trajectories depend on starting points, remain confined to a bounded region and are eventually attracted to a set of zero volume (the strange attractor with butterfly shape). 

Our goal is to evaluate the performance of two neural network architectures to predict one step ahead from each of the three time series independently as well as using all three as inputs (conditioned on history of all three series). 
For synthetic training and test data 1500 time steps are generated using numerical integration with initial values $(X0, Y0, Z0) = (0, 1, 1)$ via Euler method. The Lorenz maps are drawn using these generated time series. The values of the parameters are set in turns at $\sigma=5$, $r=20$ and $b=2$. A second evaluate setting had initial values $(X0, Y0, Z0) = (0, 1, 1.05)$ and parameters $\sigma=10$, $r=28$ and $b=8/3$, which are the settings studied in \cite{lorenz1963deterministic}. Different starting and parameter values lead to variations of the Lorenz map with the same general dynamics. A sensitivity study evaluating the short-term prediction performance to initial Lorenz starting and parameter values would be interesting but is beyond the scope of this article. Specifically, to generate the trajectories numerically, next step $x_{t+1}$ for $t+0.01$ time step for each time series uses the previous known step and the approximate time derivative for $x$, $x_{t+1} = x_{t} + 0.01\dot{x_t}$.  Then the series are rescaled to $[-0.5, 0.5]$. We then employ neural networks with the goal of learning these nonlinear relationships.

\section{Methodology}

First stop in our time travel is in 1992, to revisit work by Elsner \cite{elsner1992nonlinear}, during roughly the second wave of neural network research, or the connectionist era. In his article Nonlinear prediction, chaos and noise Elsner presents advances (at the time) in methodology for short term time series predictions based on advances in nonlinear dynamics (chaos) and parallel distributed computing. In short, the authors discuss the effectiveness of a two-layer feed forward neural network for time series prediction (one hidden layer with three neurons). For Lorenz trajectories, five time steps are used as input sequence to make a one step ahead prediction. Exactly 1000 observations are used in the train set and 500 observations in a test set. For simplicity, a validation set is not used. The predictions are made for each trajectory separately, so only in an unconditional fashion. The weights updates are made via a gradient descent with small learning rate (value is not given) and back propagation, with a mean squared error cost function. The 1992 reported RMSE is 0.072, likely averaged over the three series. 

We proceed now to evaluate more recent architectures (the third wave of neural network research) and their effectiveness in predicting the Lorenz trajectories. From the plots of the three trajectories we can see that there are some interdependencies between the time series. We will evaluate if using those relationships to make a one step ahead forecast improve the quality of our forecasts.

Our neural network training is done using the Gluon MXNet framework (\cite{chen2015mxnet}) and code is made available to exactly replicate results (up to variability inherent in neural networks solutions) on Github. Using SGD with only between 40 and 150 epochs results in training under 3 minutes per hyperparameter setting. With very minimum tuning reported results are parallel or better in some cases than the state of the art reported in the \cite{borovykh2017conditional} for Lorenz map one step ahead prediction. We model the trajectory of each of the Lorenz time series as a conditional probability of its history, as well as each other's history. In other words, the joint probability of the time series ${x_t}$ with $t = 1..T$ is modeled as 

\begin{equation}
p(x| \theta) = \prod _{t=1} ^{T} p(x_t | x_1, x_2, \dotsc , x_{t-1}; \theta)
\end{equation}

where ${x_t}$ is each of the trajectories in the Lorenz map and $\theta$ are the model parameters.

Alternatively, in the case of conditional time series one-step ahead forecast, we condition each sample at each time step on previous time steps across the multiple interrelated time series, with the model joint probability becoming

\begin{equation}
p(x|y, z, \theta) = \prod _{t=1} ^{T} p(x_t | x_1, \dotsc , x_{t-1}; y_1, \dotsc , y_{t-1}; z_1, \dotsc , z_{t-1}; \theta)
\end{equation}

The conditional probability distribution is modeled via several models to be compared: a stack of convolutional layers similar to the WaveNet architecture \cite{oord2016wavenet} and a long short term memory (LSTM) recurrent neural network \cite{hochreiter1997long}.  Comparisons are made with the feed forward neural network employed in \cite{elsner1992nonlinear}. The goal is to forecast the one step ahead, $x_{t+1}$ conditioned on $r$ previous steps of the uni or multivariate time series. This is a regression type problem, with a numerical response. 
 
\subsection{WaveNet for One Step Ahead Time Series Forecasting}

The key ingredient in the WaveNet architecture is stacking layers of dilated convolutions, in a causal fashion, so that a sample of each time step $x_t$ depends only on previous steps on not on the future, to prevent look ahead bias, as is commonplace in the time series literature and practice. More specifically, a sample at time step $t$ depends on $r$ previous time steps, where $r$ is the receptive field. The network employs 1D dilated convolutions with the purpose of learning translation invariant feature maps (what a non-dilated convolution would do) with increased receptive field, i.e., the distance in the past each sample $x_t$ sees (due to stacking dilated convolutional layers). Dilated convolutions apply a filter to the layer input skipping input values with a certain step. The output has the same size as the input (as opposed to pooling or strided convolutions). For a depiction of the a stack of dilated causal convolutional layers see \cite{oord2016wavenet}. The larger receptive field is achieved by stacking dilated convolutions. In this article, there are $L=4$ dilated convolutions layers, with $M_{l}=1$ filter of size $k=2$. Each successive layer $l$ has a dilation $d = 2^{l-1}$. By stacking the four layers, the receptive field becomes 

\begin{equation}
r = k2^{L-1} 
\end{equation}

So the sample at each time step depends on 16 previous time steps through the stack of layers. One could argue that the same dependency can be achieved by employed one larger filter, but stacking was shown to increase discriminative power and efficiency (\cite{oord2016wavenet}). The stack of dilated convolutions is preceded by a Conv1D(1x1) with no dilation and succeeded by another 1X1 convolution that outputs the one step ahead prediction. A discrete convolution between two one-dimensional signals is defined as

\begin{equation}
(f * g)(i) = \sum_{j=-\infty}^{+\infty}f(j)g(i-j)
\end{equation}

or alternatively the kernel or filter $g$ can be flipped so as we have less variation in the ranges of valid $j$ values to
\begin{equation}
(g*f)(i) = \sum_{j=-\infty}^{+\infty}f(i-j)g(j)
\end{equation}

Instead neural network software libraries typically implement the cross-correlation function
\begin{equation}
s(i) = \sum_{j=0}^{k-1}f(i+j)g(j)
\end{equation}

In our case $f$ signal should correspond to the input at each layer and $g$ filter should correspond to the convolutional kernel (filter) weights $\mathbf{w}$. For one simple example of a Conv1D layer output calculation, the cross-correlation calculation can be viewed as a multiplication of the 1D input vector by a matrix. For input vector of dimension $N=4$, $\mathbf(x) = [x_0, x_1, x_2, x_3]^{T}$ corresponding to one input example (for illustration purposes only, and ignoring bias), and a kernel of size $k=2$, $[w_0, w_1]^T$, to calculate the output feature map $\mathbf{s}$ we can form the sparse $4X3$ Toeplitz matrix:

\begin{equation}
W =
  \begin{bmatrix}
    w_0 & 0 & 0  \\
    w_1 & w_0 & 0 \\
    0  & w_1 & w_0 \\
    0 & 0 & w_1 \\
  \end{bmatrix}
\end{equation}

and the output feature map (a vector in the Conv1D case, of dimension $N_s = N - k + 1$ or the number of columns in W) is calculated via matrix multiplication as

\begin{equation}
\mathbf{s} = \mathbf{x}^TW
\end{equation}

The exposition will continue with the cross-correlation notation for consistency, with presentation of each layer. For the first Conv1D(1x1) layer the input is $(x_i)_{i=0}^{r-1}$, with $M_1=1$ filter $w$ of size $k=1$, with dilation $d=1$ (no dilation) and a bias. The feature map outputted by the first (causal, input) layer is a vector of dimension $N_1 = N-k+1 = N$,  has elements

\begin{equation}
s_1(i)= x(i)w + b_1
\end{equation}
with $i=1..N_1$.

Next we build the stack of $L=4$ dilated convolution layers. The difference between the previous convolutional feature map calculation and the dilated case is that $d$ input elements are skipped by each kernel $w$ multiplication. One can envision utilizing in an efficient manner a filter with zeros such as a $d=1$ leads to a filter (instead of filter of size $k=2$ $[w_0, w_2]^T$) like $[w_0, 0, w_1]^T$. 
Then, for each layer $l$, with dilation $d=2^{l-1}$, $M_l=1$ filters of size $k=2$, and a bias, the feature map output $s_l$ has elements

\begin{equation}
s_l(i)= \sum_{j=0}^{k-1} s_{l-1}(i-dj)w(j) + b_l
\end{equation}

where $w$ is the filter of size $k=2$. If we assume $M_{l-1}$ filters on layer $l-1$, then there are $M_{l-1}$ input channels to layer $l$ to sum over. In this case we only have one input filter. Using one filter on a layer results in learning one feature in the output feature map for that layer. The output of each layer is passed through a $ReLU(x)=max(0, x)$ nonlinearity. Furthermore, skip and residual connections are used. The skip connections are implemented as Conv1D(1x1) with dilation $d$ for each of the dilated layers and the resulting outputs for each layer $l=1..L$ are summed up and added to the residual output after layer $L$ before the final Conv1D(1x1) layer. The residual layers are implemented as $x+F(x)$ where $x$ and $F(x)$ are the input and output respectively of each of the $l=1..L$ layers. No dropout or batch normalization are included.

For the conditional case, the number of input channels on each layer changes to three, which corresponds to the number of input time series we condition on. All other architectural choices remain the same and the final Conv1D(1x1) summarizes and reshapes the output into a one step ahead forecast for the Lorenz trajectory of interest. 

To preserve causal structure where at each layer the input precedes in time the label, and to be able to start prediction at $x_0$, we pad with $r=16$ zeros that become the input for the first label $x_0$ in the training set. The input shape is a minibatch of shape batch size x number of input channels x input width.

\subsection{LSTM for One Step Ahead Time Series Forecasting}

Traditionally, parametric models have been employed for time series forecasting such as $AR$ models \cite{borovykh2017conditional}, and estimation techniques such as maximum likelihood estimation are described in standard texts such as \cite{montgomery2015introduction}. Alternative formulations in the time modeling domain include time to event prediction with similar parametric maximum likelihood based and other estimation techniques \cite{olteanu2008technical}. Time series forecasting is shown to benefit from employing neural networks if sufficient data is available \cite{borovykh2017conditional}. In the neural network framework, the most popular approach to time series prediction has been employing recurrent neural networks. As an alternative model for comparison, the Long Short Term Memory network \cite{hochreiter1997long} is employed. To maintain comparability, the sequence length (distance in the past considered to make a prediction for a time step $x_t$) is equal to the receptive field employed in the WaveNet architecture ($r=16$) and the same padding is employed. Long short term recurrent neural networks are able to retain long term information and alleviate the vanishing gradient problem (appearing in non-gated recurrent neural networks) with the addition of a few parameters (due to additional gates and memory cells) but not as many as for an equivalent feed forward neural network. We employ a one layer LSTM with 25 hidden units and a dense output layer with one neuron outputting the one step ahead prediction. We also apply a $10\%$ regularization on the hidden layer, as in \cite{fischer2018deep}. The LSTM equations are

\begin{align*}
\mathbf{i_t} = \sigma(W_{ix}\mathbf{x_t} + W_{ih}\mathbf{h_{t-1}} + \mathbf{b_i}) \\
\mathbf{f_t} = \sigma(W_{fx}\mathbf{x_t} + W_{fh}\mathbf{h_{t-1}} + \mathbf{b_f}) \\
\mathbf{o_t} = \sigma(W_{ox}\mathbf{x_t} + W_{oh} \mathbf{h_{t-1}} + \mathbf{b_o}) \\
\tilde{\mathbf{c_t}} =  \tanh(W_{cx}\mathbf{x_t} + W_{ch}\mathbf{h_{t-1}} + \mathbf{b_c})\\
\mathbf{c_t} = f_t \odot \mathbf{c_{t-1}} + i_t \odot \tilde{\mathbf{c_t}} \\
\mathbf{h_t} = o_t \odot \tanh(\mathbf{c_t}) \\
\end{align*}

The output layer dense network with one neuron outputs the one step ahead forecast taking in as input the 25 hidden activations from the LSTM's last sequence step $T$ ($T=16$). The dense layer parameters are $w_{hy}$ and $b_y$ of sizes $25x1$ and $1$ respectively. 

\begin{equation}
\hat{\mathbf{y}} = W_{hy}\mathbf{h_{T}} + b_y
\end{equation}

For each mini batch the input $X$ is shaped as sequence length x batch size X number of features. For the unconditional case, the number of features is one. For the conditional case, at each time step the number of features is three, equal to the number of time series conditioned upon. Finally, the dense layer produces the one step ahead forecast.

\section{Training and Empirical Results}

Training was done using adaptive stochastic gradient descent (Adam, \cite{kingma2014adam}) with mini batches drawn from the training set and shuffled. Training and inference took between 30 seconds and 2 minutes, depending on the model, on a 4 core Mac Pro 2.7 GHz Intel Core i7 with 16GB of memory. It was found that the SGD works better and faster than the full gradient descent optimization with 20000 iterations from In the \cite{borovykh2017conditional}. Brief hyperparameter tuning was performed, notably for learning rate, mini batch size and number of epochs selection in the absence of a validation set. 

For the LSTM architecture training, a variation of training set mini batch sampling was evaluated, where similar to natural language processing models, adjacent samples are kept next to each other in a sequential fashion and hidden states were initialized with the state at the previous batch instead of zeros. However, shuffling examples performed better. Table 1 presents the training choices for each architecture.

\begin{table}[!htbp] \centering 
 \caption{Training Choices: Unconditional Case} 
 
\begin{tabular}{@{\extracolsep{0pt}} D{.}{.}{-3} D{.}{.}{-3} D{.}{.}{-3} D{.}{.}{-3} D{.}{.}{-3} D{.}{.}{-3} D{.}{.}{-3} D{.}{.}{-3} } 
\toprule
 \multicolumn{1}{c}{Choice of Interest} & \multicolumn{1}{c}{WaveNet} & \multicolumn{1}{c}{LSTM} \\ 
\midrule
 \multicolumn{1}{c}{Number of Parameters }  & 32  & 2726 \\ 
 \multicolumn{1}{c}{Optimization Objective} & MAE  & MAE \\ 
\multicolumn{1}{c}{Optimization Algorithm} & Adam  & Adam \\ 
\multicolumn{1}{c}{Learning Rate} & 10^{-3}  & 10^{-3} \\ 
\multicolumn{1}{c}{Initialization} & He  & Xavier \\ 
\multicolumn{1}{c}{L2 regularization} & 10^{-3} & -  \\ 
 \multicolumn{1}{c}{Mini batch size} & 32 & 32 \\ 
\multicolumn{1}{c}{Number epochs} & 100 & 30 \\ 
 \multicolumn{1}{c}{Training Set Size} & 1000 & 1000 \\
 
\bottomrule

\end{tabular} 
\end{table} 

The difference in number of parameters between a WaveNet and an LSTM architecture is noticeable. The loss function is the median absolute error producing the one step ahead median estimate
\begin{equation}
J(\theta)=\frac{1}{T}\sum_{t=0}^{T-1} |\hat{x}_{t+1} - x_{t+1}|
\end{equation}
where $T$ is the size of the training set. In the convolutional neural network case the objective includes the term corresponding to the l2 penalty on the weights as well. The Xavier initialization \cite{glorot2018understanding} gives uniform initial weights in the range $[-c, c]$, where $c=\sqrt{(\frac{6}{(n_{in}+n_{out}})}$. Conversely, the He initialization adjusts to the use of rectified linear units by using a scheme of the form $Norm(0, \sqrt{(\frac{2}{n_i})})$. 

\subsection{\textbf{Multitask Learning WaveNet Architecture}}

One variation of the architecture is learning in a multitask fashion, as in Caruana \cite{caruana1995learning} and Ghosn \cite{ghosn1997multi}. Each of the three trajectories share all layers except a final Conv1D(1x1) output layer that produces a one step ahead forecast separately for each trajectory. A separate loss is defined for each output task and the gradient is based on a weighted average of the loss, with weights being tunable parameters. Training is done at the same time, as opposed to alternating batches as sometimes seen in literature \cite{dong2015multi}. Current results show improvement in test set results for $z$ trajectory prediction in this fashion, possibly due to the noise injection in training via this multitask setting. In the multitask setting the objective is

\begin{equation}
J(\theta)=\frac{1}{T}\sum_{j=0}^3 p_j\sum_{t=0}^{T-1} |\hat{s}_{t+1} - s_{t+1}|
\end{equation}

with $p_1+p_2+p_3=0$ and $ts_j$ being each one of the $x,y,z$ trajectories in the Lorenz map.

\subsection{Test Set Results}
Minimal hyperparameter tuning was performed for both the WaveNet and LSTM models, with likely significantly better results possible with more extensive tuning. As is, results are at par or better than existing benchmarks with negligible training and inference time. Predictions were made on a test set of size 500 following the train set period. A mini batch of size one was used in all scenarios. The plot displays first 100 one step ahead predictions for each of the three trajectories versus the ground truth (generated via Euler solution) for each of the three trajectories. The displayed predictions were generated from the WaveNet conditional model.

\begin{figure}[htp]

\centering
\includegraphics[width=.3\textwidth]{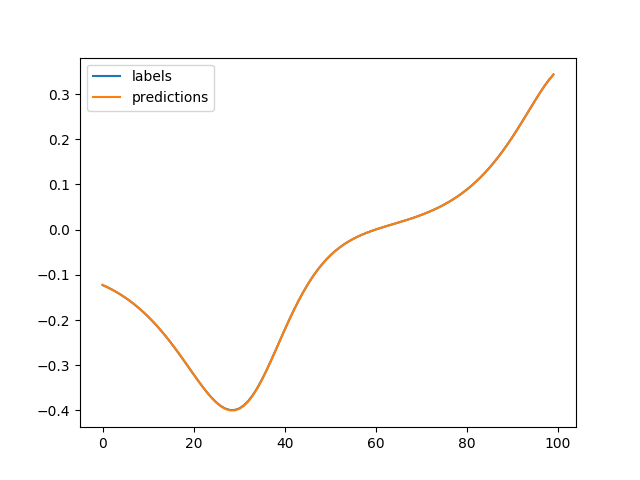}\hfill
\includegraphics[width=.3\textwidth]{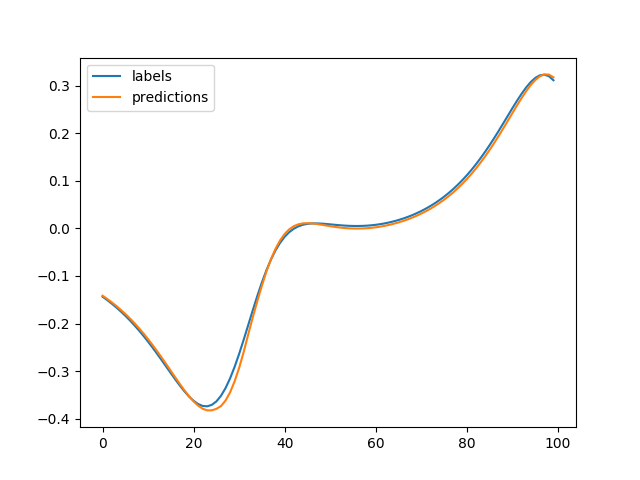}\hfill
\includegraphics[width=.3\textwidth]{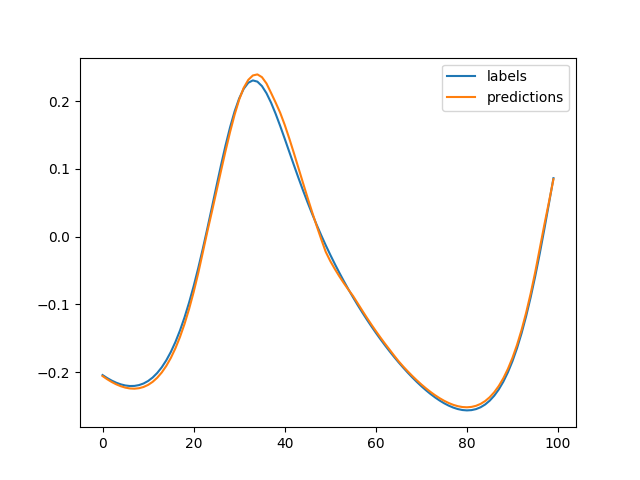}

\caption{Conditional WaveNet one step ahead forecast vs ground truth. From left to right: x, y and z trajectories respectively.}

\end{figure}

The following table presents the RMSE for a range of scenarios calculated on the test set of size 500. Several random seeds were used, for example 1234, 1235 and 42. From \cite{borovykh2017conditional} the benchmarks (average over the three series) for gradient descent with 20000 iterations unconditional WaveNet architecture is ,  with conditional WaveNet is, with LSTM is 0.00675, and from \cite{elsner1992nonlinear} with feed forward networks from 1992 the benchmark was 0.065. 

Calculated over two sets of starting values and parameters for the Lorenz map, the Conditional WaveNet produces and average ranging between 0.004 and 0.007. The LSTM architecture ranges around 0.0063. The gains in speed versus full gradient descent are at least ten fold. Even if the time series are of short history, there are gains to using a stochastic gradient descent optimization routine. In Table 2 several itemized results means and standard deviations are reported.

\begin{table}[H] \centering 
 \caption{Test set RMSE for one-step ahead forecast for x, y and z} 
\tiny
\setlength\tabcolsep{-20pt}
\begin{tabular}{@{\extracolsep{0pt}} D{.}{.}{-0} D{.}{.}{-3} D{.}{.}{-3} D{.}{.}{-3} D{.}{.}{-3} D{.}{.}{-3} D{.}{.}{-3} D{.}{.}{-3} D{.}{.}{-3}} 
\\[-1.8ex]\hline \\[-1.8ex] 
 \multicolumn{1}{c}{Model} & \multicolumn{1}{c}{x mean(std)} & \multicolumn{1}{c}{y mean(std)} & \multicolumn{1}{c}{z mean(std)}\\ 
\hline \\[-1.8ex] 
 \multicolumn{1}{c}{Unconditional WaveNet }  &   0.00480(0.00008)&  -  & - \\ 
 \multicolumn{1}{c}{Conditional WaveNet} & 0.00047(0.00020)  & 0.00750(-) & 0.00520(-)\\ 
\multicolumn{1}{c}{Multitask Conditional WaveNet} & -  & - & 0.00200(-) \\ 
\multicolumn{1}{c}{Conditional WaveNet - Lorenz parameters ($\sigma$=10, r=28, b=8/3)} & 0.00110( 0.00140)& 0.01200(0.00120) & 0.00950(0.00035)\\ 
\multicolumn{1}{c}{Unconditional LSTM random sampling} & 0.00300(0.00008)  &  0.00629(0.00043) & 0.00253(0.001)\\ 
\multicolumn{1}{c}{Unconditional LSTM random sampling: adjacent samples} & 0.01200(0.0001) & 0.00700(-) & 0.00380(-) \\ 
 \multicolumn{1}{c}{Conditional LSTM random sampling} & 0.00313(0.00068) & 0.00176(0.00017) & 0.00250(0.00014)\\ 
\multicolumn{1}{c}{Conditional LSTM - Lorenz parameters ($\sigma$=10, r=28, b=8/3)} & 0.00457(0.00045)& 0.005233(0.00130) & 0.01000(0.00082)\\ 
 \hline \\[-1.8ex] 
\end{tabular} 
\end{table} 

The suspicion that rigorous tuning can lead to significant improvement is the 10 fold improvement in x prediction RMSE over best benchmark achieved uniformly over all runs with the conditional WaveNet and different Lorenz data distributions (different $\sigma$, $r$ and $b$ parameters). The forecast for y series is the most difficult with each model. For both models, conditioning helps, the models being able to exploit nonlinear relationships between the series.

\subsection{Final Remarks}

The Lorenz map is a classical time series prediction that travelled through time as a canonical task employed throughout the neural network research history. The connectionist era results (using feed forward neural networks with one hidden layer) have been surpassed by more than 10 fold, in some cases up to 100 fold in terms of test set RMSE (using deep convolutional and recurrent neural networks) in the current era.

\bibliographystyle{plain}
\bibliography{my_bib.bib}

\end{document}